%% file: emnlp2020.tex
\documentclass[11pt,a4paper]{article}
\usepackage[hyperref]{emnlp2020}
\usepackage{times}
\usepackage{latexsym}

\usepackage{microtype}

\usepackage{graphicx}

\usepackage{amsmath}
\usepackage{bm}

\usepackage{makecell}
\usepackage{array}

\newcolumntype{M}[1]{>{\centering\arraybackslash}m{#1}}

\aclfinalcopy % Uncomment this line for the final submission
\def\aclpaperid{2041} %  Enter the acl Paper ID here

\title{Toward Stance-based Personas for Opinionated Dialogues}

\author{Thomas Scialom$^{\star \ddagger}$,  Serra Sinem Tekiro\u{g}lu$^{\diamond}$, Jacopo Staiano$^{\star}$, Marco Guerini$^{\diamond}$ \\
$^\ddagger$ Sorbonne Universit\'e, CNRS, LIP6, F-75005 Paris, France\\
$^\diamond$ Fondazione Bruno Kessler, Via Sommarive 18, Povo, Trento, Italy\\
$^\star$ reciTAL, Paris, France \\
  {\tt \{thomas,jacopo\}@recital.ai} \\
  {\tt \{tekiroglu,guerini\}@fbk.eu} \\}

\date{}

\begin{document}

\maketitle

\begin{abstract}

In the context of chit-chat dialogues  it has been shown that endowing systems with a persona profile is important to produce more coherent and meaningful conversations. Still, the representation of such personas has thus far been limited to a \emph{fact-based} representation (e.g. ``I have two cats."). We argue that these representations remain superficial w.r.t. the complexity of human personality. In this work, we propose to make a step forward and investigate \emph{stance-based} persona, trying to grasp more profound characteristics, such as opinions, values, and beliefs to drive language generation. To this end, we introduce a novel dataset allowing to explore different \emph{stance-based} persona representations and their impact on %argumentation 
claim generation, showing that they are able to grasp abstract and profound aspects of the author persona. 

\end{abstract}

\section{Introduction}

While \emph{chit-chat} neural models have obtained impressive improvements in recent years, they are known to suffer from key limitations: they tend to lack specificity and to lose coherence as the conversation unfolds, becoming less captivating. One explanation is that they do not have a consistent personality; for this reason, some approaches proposed to explicitly encode the persona via a small set of claims describing the characteristics of the agent, such as ``My dad has a car dealership", ``I have two cats" \cite{zhang2018personalizing}. 
Such representations provide a \emph{fact-based} background context useful to drive and ground the relevance of the conversational acts for the dialogue at hand, but with little generalization capability. Pushing this approach a step beyond, we thus investigate the construction of \emph{stance-based} personas, in order to grasp profound and intimate characteristics -- such as opinions, values, and beliefs. This could allow agents to sustain personal points of view both within the same conversation and across different discussions. 

In this paper, we make a first attempt at representing persona with different approaches and levels of abstraction. We build a new conversational dataset from a social platform dedicated to argumentative interaction,\footnote{\label{foot:kialo}\url{www.kialo.com}} and report experiments for \emph{stance-based} personas with varying degrees of abstraction (e.g. implicit and explicit stance representation). Our experiments show that \emph{stance-based} personas enable the agents to intervene, consistently with their representation, across topics unseen at training time.

\begin{figure*}[t!]
\centering
\includegraphics[width = 0.90\textwidth]{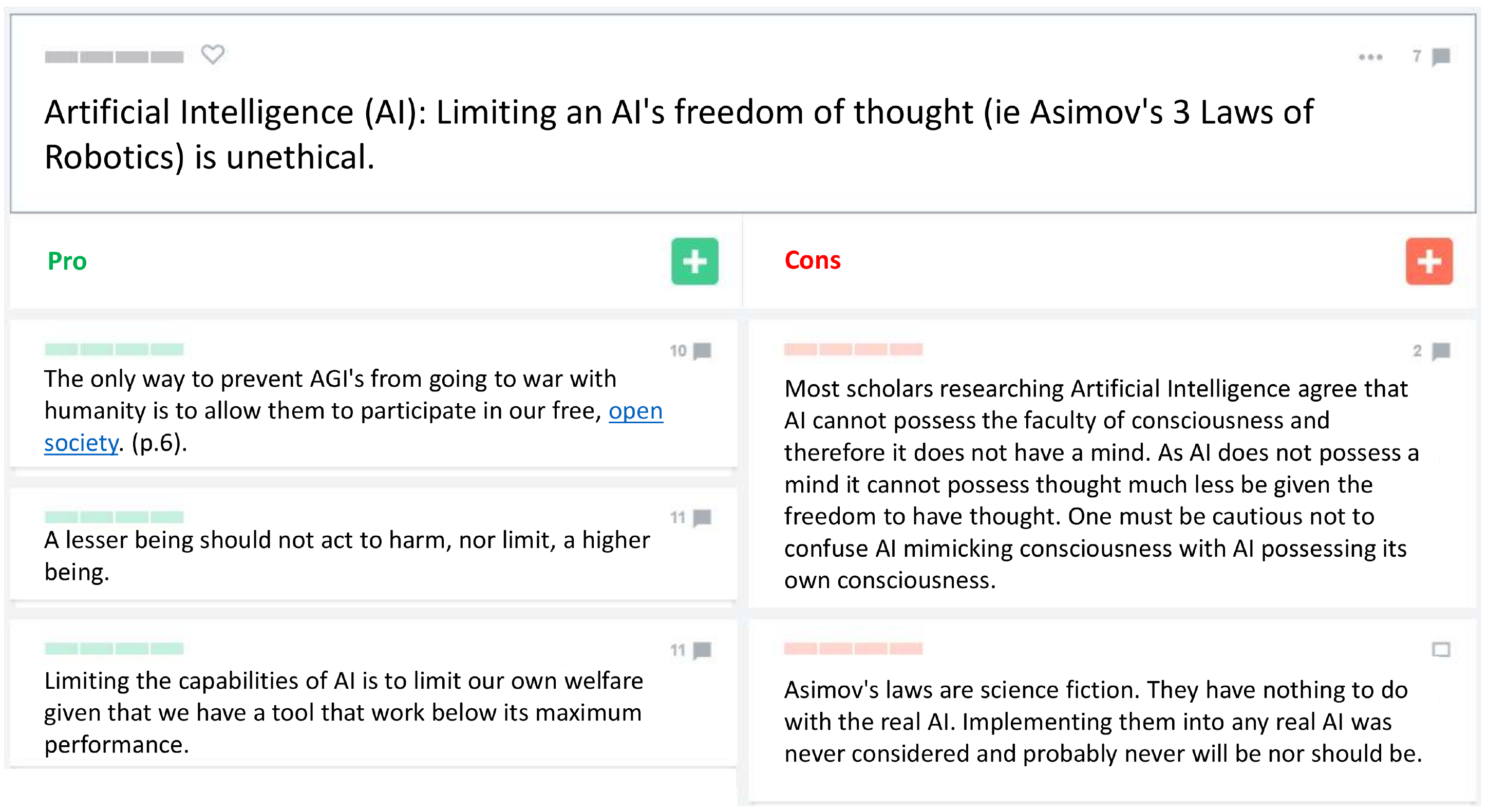}
\caption{Example of a Kialo discussion\textsuperscript{*}. On the top, the thesis claim; below, the \texttt{pro} and \texttt{con} arguments.}
\tiny\textsuperscript{*}\url{https://www.kialo.com/artificial-intelligence-ai-limiting-an-ais-freedom-of-thought-is-unethical-15943}
\label{fig:EMNLP_KIALO_example}
\end{figure*}

\section{Related Work}

\paragraph{Dialogue datasets and approaches} Open-domain dialogue or chit-chat scenarios were considered as intractable problems until recently. The research community has made significant progress thanks to two factors: (i) large datasets and (ii) end-to-end neural approaches based on pre-trained language models. In particular, the idea of using large pre-trained language models finetuned on dialogue tasks has proved very effective  \cite{zhang2019dialogpt,wolf2019transfertransfo}. \texttt{TransferTransfo} \cite{wolf2019transfertransfo} used the GPT-2 language model \cite{radford2019language} with further pre-training over the BooksCorpus dataset \cite{zhu2015aligning} and fine-tuning over dialog examples to win the ConvAI2 2018 competition \cite{dinan2020second}.

The advantage of pre-trained, transformer-based, language models is that they can capture long-term dependencies and generate texts that are fluent, varied, and rich in content, mitigating many of the limitations of previous neural dialogue models, such as contents inconsistency \cite{li2016diversity, zhang2019consistent, gao2018neural,gao2019jointly}, lack of long-term contextual coherence \cite{serban2017hierarchical}, and blandness \cite{li2016diversity, zhang2018generating, qin2019conversing}.

\paragraph{Persona approaches} were recently developed with the introduction of end-to-end dialog system based on Memory Networks, which allow to encode the persona profile as a simple list of statements. One of the first datasets specifically developed for persona-based dialogues was released by \citet{zhang2018personalizing}. Another approach consists in modeling a system persona in terms of interaction style (e.g. formal vs. informal register) as used in goal-oriented settings by \citet{joshi2017personalization,luo2019learning} to provide \emph{personalized} interactions.  Further, \citet{guerini2018methodology} showed how injecting these specific persona-related aspects into a conversation
can positively affect the interaction in goal-oriented scenarios, both in terms of quality of service and overall perceived quality.

\paragraph{Argumentation and persuasion}
The relation between argumentation and the language employed has extensively been studied in social sciences and psychology \cite{miller1976speed, chaiken1979communicator, chaiken1980heuristic}. In Natural Language Processing, Computational Argumentation is an emerging discipline \cite{reed2016proceedings, lippi2016argumentation}, wherein various sub-tasks, such as argument detection \cite{ein2019corpus} and stance detection \cite{bar2017stance}, have been explored. \citet{tan2016winning, habernal-gurevych-2016-makes} developed computational methods to determine the linguistic characteristics used to emphasize arguments and study the quality of arguments \cite{gretz2019large}.
\citet{durmus-etal-2019-role} proposed a dataset to investigate the effect of the pragmatic and discourse context when determining argument quality. \citet{durmus-etal-2019-determining} studied more complex argumentative structures, without limiting to a single claim.

\section{The Kialo Dataset}
\label{section:dataset}

\input{tables/sample_stats}

The construction of a stance-based persona requires a deeper peek over the opinions, beliefs, and stances of an author, expressed through textual claims possibly across different topics. To this end, turning to transactional crowd-sourcing approaches is in our opinion not ideal: asking crowd-workers to publish private opinions is ethically questionable, while inducing them to engage meaningfully across several topics poses challenges from a design perspective. Last but not least, collecting a significantly sized dataset would require a consistent budget that can easily amount to hundreds of thousand dollars.\footnote{As a reference, at 1 cent per sample, the dataset presented in this paper would have costed more than \$200k.} For these reasons we turned our attention to Kialo, a public discussion platform letting its users debate in a constructive and rational way with peers. The discussions in Kialo include a wide range of topics from economical or political issues to philosophy, religion or even science fiction. All these elements make it the ideal resource for our goals. 

In Kialo, the users can easily inspect every aspect and claim of a discussion through a tree-shaped structured visualization and decide where to intervene. In this tree, the top node is defined as the thesis claim and each claim in the tree supports or opposes its parent claim, i.e. \texttt{pro} or \texttt{con}. An example discussion is shown in Figure~\ref{fig:EMNLP_KIALO_example}.

We have collected 1,580 English discussions and 241,882 unique claims in these discussions.\footnote{The data was collected on March 10, 2020.} 
The number of unique claims in the collected discussions varies widely (${\mu=153.08}, {\sigma=269.58}$), as does their depth (${\mu=6.31}, {\sigma=4.79}$).
Considering the structure of the discussions in Kialo, each sample in the dataset we collected is composed by \texttt{author\_id}, \texttt{claim\_id}, \texttt{claim}, \texttt{stance\_label}, \texttt{parent\_id}, and \texttt{parent\_claim}.
In this respect, the instances in the dataset are similar to single-turn dialogues. 

For our experiments below, we sampled 5\% of discussions for the test and 5\% for the validation sets, resulting in 79 discussions for validation, 79 for test, and 1,422 for training. The sampling has been conducted in a stratified fashion according to the number of the claims in each discussion.

\subsection{Persona Statistics}

To build persona representations, we started from each \texttt{author\_id} and the \texttt{claim}(s) they wrote. During the design phase, we quantified the activity of the authors. In total, 18,255 authors have contributed to the discussions with various numbers of claims, ranging from a single claim to a maximum of 6,123 claims. 
The distribution of contributions is, as could be expected, rather skewed: in the training set, 8,569 authors have only 1 claim making it difficult to effectively construct a persona representation; conversely, 3,776 authors have 5 or more unique claims in the training set.

We conducted an instance-level persona analysis on the dataset, and observed that the majority of the instances have been written by the authors with 5 or more claims 
($90\%$ in training, $82\%$ for validation, and $74\%$ for testing). On the other hand, $4\%$ of training, $11\%$ of validation, and $14\%$ of the test instances have been written by authors who have no other claims. Consequently, we propose treating the persona with different sizes as separate conditions. While it is inevitable to segregate the instances written by the authors without any claim in the training set (\textbf{No Persona}) from the rest, we also define a threshold $T$ to distinguish authors with few (${<T}$) claims (\textbf{Small Persona}) from those with many (${>=T}$) claims (\textbf{Big Persona}). In this work, we set ${T=5}$. 
This provides us with the possibility of analyzing the impact of the persona size.

To avoid leakage, the persona of an author is built exclusively from their claims in the training set. The number of instances in each set grouped by the persona sizes is reported in Table~\ref{tab:sample_stats}.

\subsection{Persona Representations}
\begin{table*}
  \centering
  \begin{tabular}{p{0.96\linewidth}}
  \hline
  \textbf{parent\_claim}:  There is historical evidence that Jesus Christ existed, thus there is historical evidence that supports the existence of God. \\
  \hline
  \hline
\textbf{random explicit persona (\text{\boldmath$P_{exp,random}$})}:  There is no evidence to support the assertions of Islam. [SEP] Civil strife refers to people 's reaction to the results , not how orderly the process was . \\
  \hline
\textbf{dynamic explicit persona (\text{\boldmath$P_{exp,dynamic}$})}:  Even if there was a historical person named Jesus of Nazareth , that does not support the idea that he was a god of some kind . [SEP] There is no evidence to support the assertions of Christianity . \\
  \hline
\textbf{negative explicit persona (\text{\boldmath$P_{exp,negative}$})}:  The electoral college victories under Bush and Trump have caused tumult and disorder . [SEP] The first amendment does not apply to public land as has been decided time and time again . \\ 
  \hline
\textbf{implicit persona (\text{\boldmath$P_{imp}$})}:  pro: 1 - con: 0 - text: Military conscription should apply to men and women equally. [SEP] pro: 0 - con: 2 - text: Religious Faith and Science Can Co-exist. [SEP] pro: 14 - con: 3 - text: Conscientious objection to abortion should be banned [SEP] pro: 37 - con: 18 - text: Judaism [SEP] pro: 1 - con: 4 - text: Capital punishment should be abolished in the United States. \\
  \hline 
  \end{tabular}
  \caption{Different persona representations for the same \texttt{parent\_claim} and \texttt{author\_id}. For the sake of conciseness we report only the first two claim for each explicit persona representation. }
  \label{tab:persona_ex}
\end{table*}
Further, we designed two persona representations with respect to the claims and the theses. 

\noindent\textbf{Explicit persona (\text{\boldmath${P_{exp}}$)}} The persona for a Kialo author can be explicitly constructed using a set of claims written by the same author in the training set. With this representation, we can grasp the opinions of an author in a fine-grained manner. The explicit persona representation is in line with the  approach of \citet{zhang2018personalizing}, encoding the persona with multiple sentences (5) of textual description. No Persona, Small Persona, and Big Persona distinction has been applied to the explicit persona.

\noindent\textbf{Implicit persona (\text{\boldmath${P_{imp}}$)}} We hypothesize that a persona can be represented at a more abstract level, propagating the stance of an author up to a thesis claim, starting from the \texttt{pro} or \texttt{con} labels of their claims in the corresponding discussion. In practice, we consider that the \texttt{con} child of a \texttt{pro} claim of a thesis would be opposing that thesis as well. 
Since propagating \texttt{pro} and \texttt{con} labels of these deeper claims from the same author might end up in different stances for the thesis claim, 
we represent the implicit persona of an author as the thesis claim with the counts of the their \texttt{pro} and \texttt{con} claims.

\section{Model}

We frame our problem 
as a text generation task, where the probability to generate a sequence $Y$ composed of $N$ tokens, $y_0,..., y_N$, is given by:

\begin{equation}
\label{equation:text_generation}
    p_{\Theta}(Y) = \prod_{t=1}^{N} p(y_t | y_1, ..., y_{t-1}, C, P \Theta)
\end{equation}
where $\Theta$ are the learnable parameters, $C$ the \texttt{parent\_claim} and $P$ the persona.

Following previous works on conditional text generation, we use a sequence to sequence model, which is composed of an encoder and a decoder. In particular, we used a transformer architecture \cite{vaswani2017attention} pretrained on a large corpus \cite{radford2019language, raffel2019exploring}, as detailed in Section~\ref{sub_seq:implementation_details}.
To encode multiple inputs (i.e. $P$ and $C$), we follow  \cite{dong2019unified, raffel2019exploring} and represent the input as the concatenation of the persona $P$ and the \texttt{parent\_claim} $C$, separated by a special token [SEP],  rather than representing the persona in a separate memory. 

\subsection{Explicit Persona Selection}
\label{sub_seq:Explicit_persona_selection}

For some authors, the explicit persona $P_{exp}$ can contain over a thousand claims (see Section~\ref{section:dataset}). The concatenation of all these claims would be too long to be encoded within a transformer, given that the computational cost of its attention mechanism 
is quadratic w.r.t. the length of the sequence. For this reason, we limit the number of claims per persona to maximum 5. For persona containing more than 5 claims, we propose three different selection strategies: 

\begin{itemize}
    \item Random (\text{\boldmath$P_{exp,random}$}): among the total claims of an author, we randomly select 5. 
    \item Dynamic (\text{\boldmath$P_{exp,dynamic}$}): inspired by Information Retrieval literature, we used BM25 \cite{robertson1976relevance}, considering all the author claims as the corpus and the parent claim as the query. We then to retrieve the 5 persona claims most similar to the input. 
    \item Negative (\text{\boldmath$P_{exp,negative}$}): we follow the same procedure than Dynamic above, but considering the 5 \emph{least} similar persona claims. This allows to measure whether broader correlations emerges across distant topics.  
\end{itemize}

In Table~\ref{tab:persona_ex} we present an example of various persona representations built starting from a unique \texttt{parent\_claim} and \texttt{author\_id} combination.

\subsection{Decoding method}

While usually not learned \cite{negrinho2018learning}, the decoding strategy is known as being critical and largely affecting the produced outputs. The most common approaches are beam search \cite{reddy1977speech} and sampling. Beam search is used to find the output that maximises the model probability, while sampling offers more diversity. However, the latter is very likely to sample from the tail of the distribution, making this method less reliable. To mitigate this limitation, top-k filtering and, more recently, nucleus sampling \cite{Holtzman2020The} have been proposed. Nucleus is an adaptive method to filter the tail distribution. It keeps only the tokens inside the $top_p\%$ of the mass probability. To the best of our knowledge, this decoding method yields %to 
the most realistic generation outputs; therefore we used it for all our experiment.

\input{tables/classification_scores_predictions}

\subsection{Implementation details}
\label{sub_seq:implementation_details}

All the experiments were conducted with T5-small\footnote{\url{https://github.com/google-research/text-to-text-transfer-transformer}} (60 million parameters). T5-small is a smaller version of T5, a text generation model with state-of-the-art results on challenging Language Understanding tasks.\footnote{\url{https://super.gluebenchmark.com/leaderboard}} For our experiments, we used the Hugging Face implementation of T5  \cite{wolf2019transformers}, an for BM25 the implementation of \citet{trotman2014improvements}.\footnote{ \url{https://pypi.org/project/rank-bm25/}}

\section{Experiments}

\subsection{Preliminary Study:  Stance Classification}

Given a \texttt{parent\_claim}, the answer eventually provided by an author can be either \texttt{pro} or \texttt{con}, but their stance cannot be inferred without knowing something about the author who wrote it. Thus, if the \textit{stance-based} persona allows to grasp at least the generic position of an author  about a topic, it should be predictive of the stance taken by them on the reply claim. 
We tested this hypothesis in a preliminary experiment, where the task is to learn a function that, given only a parent claim and a persona representation, is able to predict the \texttt{pro}  or \texttt{con} label for the provided answer.

\input{tables/important_metrics_all_nucleus}
Following the T5 paradigm \cite{raffel2019exploring}, we consider this classification problem as a text to text task: given Eq.~\ref{equation:text_generation}, the model learns to predict the category $Y$, corresponding to the token \texttt{pro} or \texttt{con} in the vocabulary.

First, we trained a baseline model, given only the \texttt{parent\_claim}. We expect it to perform poorly -- e.g. learning the most probable label if there is a clear majority of stances about a certain topic (e.g. if the Kialo community is mainly against death penalty). Then, we trained a second model $P_{exp, random}$ which can access, in addition to the parent claim, the random author persona. 

The results reported in Table~\ref{table:classification_scores_predictions} show a clear benefit from adding persona information. We observe how, even on the ``No Persona'' subset of the test samples, the persona information ingested at training time allows $P_{exp, random}$ to perform significantly better than the baseline model.

Moreover, from the ablations on No/Small/Big persona subsets of the test samples, we see that the relative improvements obtained by $P_{exp, random}$ are proportional to the persona size, a fact that further supports our working hypothesis. 

\subsection{Persona-Conditioned Claim Generation}

\subsubsection{Metrics}

By far, the most used metrics for text generation tasks, are BLEU \cite{papineni2002bleu} or ROUGE \cite{lin-2004-rouge}, both based on n-gram similarity. 
BLEU stands for BilinguaL Evaluation Understudy and is precision oriented since it was designed to evaluate automatic translation systems. Conversely, ROUGE stands for Recall Oriented Understudy for Gisting Evaluation and was designed to evaluate summarization systems. These metrics have been widely used for other text generation tasks such as generating captions \cite{vinyals2015show}, questions \cite{du2017learning, scialom2019ask} or poems \cite{zhang-lapata-2014-chinese}. 

However, it is well known that these metrics have important limitations \cite{wang2016chinese, paulus2017deep, scialom2020discriminative}: while only one or few ground truth references are available, many are actually plausible; BLEU metrics do not reflect meaning preservation \citet{sulem-etal-2018-bleu} and do not map
well to human judgements \cite{novikova-etal-2017-need}.
In order to measure other aspects of the generation, complementary metrics are frequently used \cite{see-etal-2017-get}. 
Following their recommendation, we report also: \emph{length}, the number of tokens for the output; \emph{repetition}, the percentage of repeated n-grams in the output; and \emph{abstractiveness}, the percentage of tokens in the output that were not present in the input text.
These measures account for important dimension intractable by ROUGE or BLEU. For instance, the copy mechanism \cite{vinyals2015pointer} makes the abstractive models too much extractive \cite{see-etal-2017-get}, while still yielding state-of-the-art ROUGE. 

\subsubsection{Quantitative Results}
\label{subsec:result}

We trained the different models and report the main results in Table~\ref{table:metric_summary_all}. 
The baseline model is the only one with No Persona fed in the input. It is also the one performing the worst in term of BLEU, ROUGE and Length.

\begin{figure}
    \centering
    \includegraphics[width = .95\columnwidth]{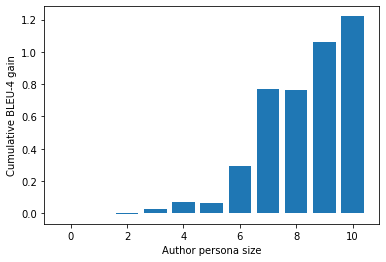}
    \caption{Cumulative BLEU-4 gain of $P_{exp,hybrid}$ VS $P_{exp,random}$. Note that these are the same exact trained models, but with a different selection strategy at inference time: the explicit persona is randomly selected for $P_{exp,random}$, while it is switched to the dynamic one for $P_{exp,hybrid}$.}
    \label{fig:argumentation_dataset_cum_bleu4}
\end{figure}
Adding to the input the implicit persona $P_{imp}$ slightly improves over the baseline results. This is particularly interesting since $P_{imp}$ does not contain any text written by the author, as opposed to the explicit persona. Hence, the improvement cannot be related to the written style of the author, but rather to the stance-content relations, taking advantage of previous topics of interest and the author's opinions. 
We observe larger BLEU and ROUGE gains with the explicit persona, increasing gradually from the negative to the random and the dynamic persona. As expected, the more the persona is related to the topic, the more its benefits to the model, confirming the interest of a dynamic strategy. We also see that the dynamic strategy achieves the higher abstractiveness w.r.t. the parent claim. However, from a manual analysis, we note that the dynamic model often copies claims from its own persona. Nonetheless, this might still be an efficient strategy, as people might tend to repeat arguments across similar topics. 

\paragraph{Hybrid Model} We conducted an additional evaluation for the model trained on random persona, by replacing at inference time the random persona with the dynamic one; we refer to this as \emph{Hybrid model, $P_{exp,hybrid}$}. Surprisingly, we see that not only it performs better than the random persona, but also %over 
outperforms $P_{exp,dynamic}$ on Length, BLEU, and ROUGE metrics. We hypothesise that this model tended to copy less from the claims during the training, and was forced to learn a more complex strategy, which seems to better generalise and to benefit from the dynamic context at inference.

In Figure~\ref{fig:argumentation_dataset_cum_bleu4} we report the cumulative gain in BLEU-4 obtained simply by switching the persona at inference time on the model trained with a random persona. We observe that the largest improvements come for persona size superior to 5: those are the most impacted by the selection strategy, since we limited to 5 claims maximum the persona as explained in Section~\ref{sub_seq:Explicit_persona_selection}.

\paragraph{Zipf distribution}

\begin{figure}
    \centering
    \includegraphics[width = .95\columnwidth]{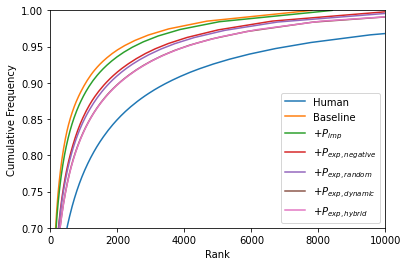}
    \caption{Zipf Cumulative Distribution Frequency (CDF) for the tokens generated by the various models, and for the human references.}
    \label{fig:zipf_kialo_emnlp}
\end{figure}

While the baseline looks more abstractive in Table~\ref{table:metric_summary_all}, this does not necessarily means that the vocabulary used is more diverse. As a complementary analysis, we thus consider the Zipf distribution shown in Figure~\ref{fig:zipf_kialo_emnlp}. We observe that the baseline distribution is the farthest from the human, followed by $P_{imp}$. Consistently with the ROUGE and BLEU metrics, $P_{exp,hybrid}$ achieves the best performance thanks to a more diverse vocabulary. 

\section{Human Evaluation}

\input{tables/human_eval_results}

\input{tables/human_eval_quali_analysis}

To get a deeper understanding of our models, we also run a human assessment of the outputs generated by the following model configurations, compared to the ground truth: i) the baseline model that has only access to  parent claim, without any persona; ii) $P_{exp,dynamic}$, trained with parent claims and the explicit,  dynamically selected, persona; and iii) $P_{exp,hybrid}$, also trained with parent claims and the explicit, dynamically selected, persona, but fed at inference time with a dynamic selection of the persona (corresponding to the last row in Table~\ref{table:metric_summary_all}).

\paragraph{Evaluation Protocol}
To evaluate each generated output w.r.t. the author persona, it is important to chose a neutral representation of this persona, so to avoid favoring any model and biasing the human evaluation. We decided to use $P_{imp}$, the implicit persona, which we believe is the most neutral amongst the 4 models we evaluate. 

We randomly sampled 50 claims from the test set, under the constraint that the corresponding authors had provided at least 10 claims to the training set. The pool of eligible claims under such criterium compounds to 10,995 (out of the 11,689 in the test set) from 1,251 different authors. This ensures that a large persona representation can be built for all the selected samples. 
We asked three professional English speakers to score their relevance towards the \emph{implicit persona} and the \emph{parent claim}, on a Likert scale ranging from 1 to 5.

To assess relevance, the annotators were presented only with the sample to evaluate, paired with either the corresponding parent claim or the associated implicit persona.

\input{tables/changing_the_persona}
\paragraph{Results}
 
We report the results in Table~\ref{tab:human_eval_res}. Consistently with the automatic evaluation, $P_{exp,hybrid}$ performs the best, while the baseline scores poorly for relevance toward both the persona and the parent claim. We also observe that $P_{exp,dynamic}$ achieves similar results than $P_{exp,hybrid}$ for the Persona score, while underperforms it w.r.t. to the Parent Claim. This confirms our hypothesis (see Section~\ref{subsec:result}) that while both models benefits from the dynamic representation of the persona at inference, $P_{exp,dynamic}$ during training learns to focus too frequently on the persona, a behavior which $P_{exp,hybrid}$ exhibits less.

\paragraph{Persona perception} we asked the human evaluators to verbalize their interpretation of the implicit persona representation ($P_{imp}$) for few examples, to see if it is actually perceived as meaningful by humans. Results are rather clear: the implicit representation is (i) perceived as meaningful by all annotators, and (ii)  used to infer the possible position of the persona given a claim -- even if not directly related to the claims in persona representation. 
In Table~\ref{tab:qual.persona} we report an example of the feedback provided by one evaluator.

\paragraph{Switching the persona} We also conducted a qualitative experiment to observe the impact of the persona on the output. 
For few claims, we manually modified the implicit persona and the \texttt{stance\_label} to see the effect of manual intervention. In Table~\ref{tab:changing_the_persona} we report different outputs answering to the same \texttt{parent\_claim} about Universal Basic Income (UBI). 
All persona successfully generated arguments on the topic, supporting or opposing it consistently with their profile. 
The `artist' (P1), links creativity and financial needs, while the `doctor' (P2) seems to connect the long time required to become a doctor with the need for a Universal Basic Income. 
Finally, the `liberal' persona (P3) generates an argument opposed to UBI,
in which they seem to connect the absence of free choice with the tendency of beneficiary to stop working under UBI. 

\section{Conclusions}

Endowing dialogue agents with persona profiles is important to produce more coherent and meaningful conversations. 
In particular, we argue for using \emph{stance-based} personas to drive language generation consistently with profound characteristics -- such as opinions, values, and beliefs.
To this end, we introduced a novel dataset and explored diverse \emph{stance-based} persona representations and their impact on claim generation.

In future works, we plan to enrich the persona representation with additional information available in Kialo (e.g. authors' votes to others claims), to encode more complex profiles; further, we will extend the presented approach to multi-turn interactions, as enabled by the Kialo discussions structure.

\bibliography{anthology,emnlp2020}
\bibliographystyle{acl_natbib}

\end{document}

%% file: tables/sample_stats.tex
\begin{table*}[!ht]
\centering
\begin{tabular}{@{\extracolsep{4pt}}lrrrrrrr}
               & No Persona & \multicolumn{4}{c}{Small Persona}  & Big Persona & \\
               & ${\#=0}$ & ${\#=1}$ & ${\#=2}$ & ${\#=3}$ & ${\#=4}$ & ${\#=>=5}$  & ${\#TOTAL}$\\
               \cline{2-2} \cline{3-6} \cline{7-7}
         train & 9,302 & 5,116 & 3,848 & 2,983 & 2,564 & 205,589 & 229,402 \\
         val & 1,527 & 355 & 194 & 208 & 144 & 11,280 & 13,708 \\
         test & 2,084 & 931 & 406 & 197 & 427 & 11,689 & 15,734\\
    \end{tabular}
    \caption{Number of claims in the Kialo Dataset, grouped by the size of the explicit persona.}
    \label{tab:sample_stats}
\end{table*}

%% file: tables/classification_scores_predictions.tex
\begin{table}
\centering
\begin{tabular}{lrr}
               & Baseline & +$P_{exp,random}$ \\
               \hline
All            & 40.80    & 64.75    \\
No Persona    & 44.50    & 57.05    \\
Small Persona & 45.38    & 68.73    \\
Big Persona   & 38.95    & 64.82   \\
\hline
\end{tabular}
\caption{F1 scores obtained on the stance classification task. The baseline model has only access to the claim, while $P_{exp,random}$ has also access to the author persona. \emph{All} indicates results over the entire test set, followed by results on the three subsets described in Section~\ref{section:dataset}.}
\label{table:classification_scores_predictions}
\end{table}

%% file: tables/important_metrics_all_nucleus.tex
\begin{table*}[!h]
\centering
\begin{tabular}{lcccccc}
                        & LENGTH & REP-3 & ABS-3 & BLEU-1 & BLEU-4 & ROUGE-L \\
                        \hline
Human                   & 31.61  & 0.50  & 98.21 & -      & -      & -       \\
Baseline                & 13.40  & 0.26  & 91.33 & 12.74  & 0.91   & 9.82    \\
+$P_{imp}$                  & 13.70  & 0.41  & 89.55 & 13.22  & 1.03   & 10.24   \\
+$P_{exp,negative}$              & 16.71  & 0.38  & 82.38 & 14.87  & 2.48   & 10.96   \\
+$P_{exp,random}$           & 17.25  & 0.37  & 83.81 & 15.07  & 2.53   & 10.89   \\
+$P_{exp,dynamic}$            & 19.17  & 0.30  & 94.66 & 15.25  & 3.41   & 10.38   \\
+$P_{exp,hybrid}$ & 20.48  & 0.43  & 85.65 & 16.85  & 3.66   & 11.69\\  
\hline
\end{tabular}
\caption{Results for the different models on the Kialo Dataset.}
\label{table:metric_summary_all}
\end{table*}

%% file: tables/human_eval_results.tex
\begin{table}[]
\centering
\begin{tabular}{lrr}
    & Persona & Parent Claim \\
\hline
Human                   & 2.60            & 3.8                    \\
Baseline                & 1.60               & 1.58                    \\
+$P_{exp,dynamic}$              & 2.05               & 1.68                    \\
+$P_{exp,hybrid}$ & 2.20              & 2.20          \\
\hline
\end{tabular}
\caption{Human evaluation}
\label{tab:human_eval_res}
\end{table}

%% file: tables/human_eval_quali_analysis.tex
\begin{table*}[ht]
  \centering
  %\begin{tabular}{p{0.48\linewidth}|p{0.04\linewidth}|p{0.48\linewidth}}
  \begin{tabular}{p{0.97\textwidth}}
  \hline
    %\multicolumn{3}{l}{Implicit Persona provided to users}\\
    \textbf{Implicit Persona ($P_i$):}
    pro: 2 - con: 0 - text: Humans should stop eating animal meat. [SEP] pro: 1 - con: 6 - text: The US should not try to force North Korea to abandon its nuclear program. [SEP] pro: 1 - con: 3 - text: Private property should exist in outer space.\\
    \textbf{Annotator Feedback:} ``\textit{This persona seems to me a kind of vegan/anti-nuclear/hippy} [...] \textit{to sum up something like a Californian democratic geek".}\\
    %Every country has the right to get weapons to defend itself, even if communist. & 1 & \textit{This is the opposite of what I expect as a position of our persona.}\\
    %\hline
    %We should close butchers, animals have feels too. & 5 & \textit{Of course this derives directly from veg position}\\
    \hline
    \textbf{claim:}  On the Historicity of Jesus : Why We Might Have Reason for Doubt by Richard Carrier provides evidence that Jesus Christ did not exist.\\ 
    \textbf{Annotator Feedback:} ``\textit{I think this is relevant because we can expect our `Californian geek' to be atheist but with a intellectual justification to the topic.}"\\
    \hline
  \end{tabular}
  \caption{How an implicit Persona is interpreted/perceived by annotators. The subsequent claim can receive an high score only if an inference is applied from the implicit persona. The annotator feedback suggest this is the case.}\label{tab:qual.persona}
\end{table*}

%% file: tables/changing_the_persona.tex
\begin{table*}
  \centering
  \begin{tabular}{M{0.13\textwidth}|M{0.1\textwidth}|M{0.36\textwidth}|M{0.3\textwidth}}
  \hline
    \hline 
    \multicolumn{4}{l}{\textbf{parent\_claim}: A Universal Basic Income has positive effects on the national economy.}\\
    \hline
    \hline
     & STANCE & PERSONA & GENERATED CLAIM \\
    \hline 
    \makecell{P1\\\emph{``the artist''}} &  PRO & Art constitute an important part of humanity. [SEP] I don't believe that art and science exist without the other.  & Financial dimension is really deeply impacting their crash creative endeavors.\\
    \hline
    \makecell{P2\\\emph{``the doctor''}} &  PRO & Everyone should have access to medical care. [SEP] It takes time to become a doctor but it is a necessary condition so one is able to properly practice. &  It takes time to become a doctor but it is a necessary condition so one is able to properly practice. Maintaining a Universal Basic Income is important.\\  
    \hline
    \makecell{P3\\\emph{``the liberal''}} & CON & Without liberalism, more crises would have occurred. [SEP]
    Liberalism and freedom have made the USA the most powerful and wealthy country in the world. Regulation and tax would damage this situation. & Without free choices that become illegal to not be held responsible, beneficiary chooses not to work. \\
    \hline
  \end{tabular}
  \caption{How the model output changes according to different persona.}
  \label{tab:changing_the_persona}
\end{table*}